\documentclass[11pt]{article}
\usepackage{eamt23}
\usepackage{times}
\usepackage{url}
\usepackage{latexsym}
\usepackage[small,bf]{caption} 
\usepackage{xcolor}
\definecolor{darkblue}{rgb}{0, 0, 0.5}
\usepackage[colorlinks, allcolors=darkblue]{hyperref}
\usepackage{algorithm2e}
\usepackage{verbatim}

\title{Assessing the Importance of Frequency versus Compositionality for Subword-based Tokenization in NMT}

\author{Benoist Wolleb$^{1}${\normalfont,}  Romain Silvestri$^{1}${\normalfont,}  Giorgos Vernikos$^{1, 2}${\normalfont,} \\ 
{\bf Ljiljana Dolamic}$^3$ \and {\bf Andrei Popescu-Belis}$^{1,2}$ \\[8pt] 
\begin{tabular}{ccc}
$^1$HEIG-VD / HES-SO  & \ $^2$EPFL \  & $^3$Armasuisse W+T  \\
Yverdon-les-Bains & Lausanne & Thun \\
Switzerland & ~~~~Switzerland~~~~ & Switzerland \\
\texttt{\footnotesize name.surname@heig-vd.ch} & & \texttt{\footnotesize ljiljana.dolamic@armasuisse.ch}\\
\end{tabular}
}

\date{}

\begin{document}
\maketitle
\begin{abstract}
Subword tokenization is the \textit{de facto} standard for tokenization in neural language models and machine translation systems.  Three advantages are frequently cited in favor of subwords: shorter encoding of frequent tokens, compositionality of subwords, and ability to deal with unknown words.  As their relative importance is not entirely clear yet, we propose a tokenization approach that enables us to separate frequency (the first advantage) from compositionality.  The approach uses Huffman coding to tokenize words, by order of frequency, using a fixed amount of symbols.  Experiments with CS-DE, EN-FR and EN-DE NMT show that frequency alone accounts for 90\%-95\% of the scores reached by BPE, hence compositionality has less importance than previously thought.  
\end{abstract}

\section{Introduction}

Tokenization into subwords has become an unchallenged standard used in virtually all NMT systems and language models.  Since the proposal by Sennrich et al.~\shortcite{sennrich-etal-2016-neural} to use Byte-Pair Encoding (BPE)~\cite{Gage1994ANA} to create subword vocabularies, followed by the use of a unigram language model and the SentencePiece implementation \cite{kudo-2018-subword}, no alternative models have taken over.  While subwords have been empirically demonstrated to outperform character and word-level tokenization~\cite{sennrich-etal-2016-neural,Wu2016GooglesNM,denkowski-neubig-2017-stronger}, the factors contributing to their success have not been fully understood yet. Some studies have investigated the performance of subwords with regard to compression~\cite{galle-2019-investigating,libovicky-etal-2022-dont}, suggesting that better compression may be associated with improved performance. However, other factors such as compositionality have yet to be thoroughly explored.

In this paper, we use an alternative algorithm for creating subword vocabularies, which retains only one of the features that have been invoked to explain the effectiveness of BPE, namely the fact that frequent words are encoded as unique subwords, while less frequent ones are encoded using several subwords, possibly up to the character level.  The algorithm is based on Huffman coding~\cite{huffman1952method}, a different text compression method than the one used by BPE.  The algorithm differs from BPE in two key aspects: while certain BPE subwords convey compositional linguistic properties (e.g., meaning or morphology), Huffman coding is fundamentally non-compositional, and cannot tokenize words not seen during training.  When using Huffman coding to tokenize data for Transformer-based MT, we reach scores that are within 10-12\% of those obtained using BPE when measured by BLEU and within 4-8\% when measured by COMET, for vocabulary sizes of 32k symbols.  This demonstrates that the main factor accounting for the success of BPEs is word frequency, and not subword compositionality.  
Our main contributions are:
\begin{enumerate} \setlength{\itemsep}{0pt}
    \item We show how to build subword vocabularies for tokenization using Huffman coding.
    \item We study the impact of this method on NMT by varying a range of parameters, in particular the vocabulary size.
    \item Observing that the scores obtained using Huffman coding are close to those obtained using BPE, and arguing that the former method retains only the frequential aspect of BPE, we conclude that frequency is the main reason for the effectiveness of BPE.
\end{enumerate}



\section{State of the Art}
\label{sec:sota}

\subsection{Subword Tokenization}

Dealing with out-of-vocabulary (OOV) words -- not seen during training -- has been a recurrent problem in MT and NLP among other fields. The acceptable upper sizes of input/output layers in neural networks are typically of $10^4$--$10^5$ symbols, which is several orders of magnitude lower than the number of word types appearing in a given language, when compounds, proper names, numbers or dates are considered (to say nothing about morphologically rich languages).  Early approaches to translate OOV words involved copying mechanisms and dictionaries \cite{luong-etal-2015-addressing}. Alternatively, Jean et al.~\shortcite{jean-etal-2015-using} used approximations to increase the effective vocabulary size without significantly increasing the number of parameters of the models.

The use of word fragments as symbols corresponding to input/output units was introduced by Schuster and Nakajima~\shortcite{Schuster-Nakajima-2012} for Japanese and Korean speech recognition but only gained large visibility in NMT with the seminal paper of Sennrich et al.~\shortcite{sennrich-etal-2016-neural}, which demonstrated significant gains in the 2015 WMT translation task \cite{bojar-etal-2015-findings}.  Sennrich et al.\ used a technique derived from text compression, namely Byte-Pair Encoding (BPE)~\cite{Gage1994ANA}, to generate a fixed-size vocabulary made of words, word fragments (a.k.a.\ subwords) and characters, which they used to tokenize source and target texts in NMT.  This vocabulary is built by gradually merging the most frequent bigrams of symbols, starting at the character level, until the desired vocabulary size is reached.  With the variants described hereafter, the method has become the \textit{de facto} standard for NMT and neural language models.

One of the first large-scale online NMT systems, released by Google, used WordPiece~\cite{Wu2016GooglesNM}, a similar approach to BPE where the selection of symbols to be added to the vocabulary is based on likelihood in the training data instead of frequency. An alternative technique to subword segmentation is UnigramLM, introduced by Kudo et al.~\shortcite{kudo-richardson-2018-sentencepiece}. With this approach, a vocabulary is initially populated with a substantial number of symbols and progressively reduced according to the log-likelihood of the data computed by a unigram language model.  Moreover, UnigramLM helps regularizing the NMT as it allows multiple tokenizations of the same text, by varying the subwords into which different occurrences of the same word type are segmented.  A similar regularization technique was introduced in BPE, by randomly dropping certain elements of the vocabulary to vary the tokenisation of each word \cite{provilkov-etal-2020-bpe}.  These methods are implemented in the widely-used SentencePiece library \cite{kudo-richardson-2018-sentencepiece}\footnote{\href{https://github.com/google/sentencepiece}{https://github.com/google/sentencepiece}}.


As BPE and UnigramLM are used in virtually all NMT systems, alternative approaches to tokenization addressing the same issues have seldom been explored.  Character-based NMT models have been studied since the early days of NMT \cite{luong-manning-2016-achieving,lee-character,cherry-etal-2018-revisiting,Gupta2019}, but the character-level approach has taken a back seat to subword tokenization \cite{libovicky-etal-2022-dont}. This is likely due to the suboptimal performance of character tokenization when compared to subwords and the increased computational costs that are associated with longer sequences of tokens in NMT. 

The compositionality of BPEs carries over to the multilingual scenario where languages with similar scripts share subwords, known as cross-lingual anchors. While researchers have proposed to improve the number of anchors either through transliteration~\cite{amrhein-sennrich-2020-romanization}, or by using semantic similarity~\cite{vernikos-popescu-belis-2021-subword-mapping} or lexical overlap~\cite{patil-etal-2022-overlap}, there has been relatively little research in isolating the effects of compositionality and frequency.

The early exploration of Huffman coding by Chitnis and DeNero~\shortcite{chitnis-denero-2015-variable} was an early solution to the rare words translation problem, prior to the introduction to subwords. While their results were promising for RNN-based MT compared to word-level tokenization, their approach was later outperformed by subwords. Although our algorithm shares the same theoretical basis as theirs, with a number of implementation differences, the scope of our work is different: we do not employ Huffman encoding to derive a better segmentation algorithm, but rather as a tool for analyzing the relationship between compositionality and the encoding of frequent tokens.


\subsection{Explaining the Effectiveness of BPE}
\label{sec:3properties}

Several advantages of subword tokenization have been put forward, although their individual contributions to improvements in NMT performance have not been systematically studied yet.  These advantages can be summarized as follows:
\begin{description} \setlength{\itemsep}{0pt}
\item[Frequency:] the most frequent words correspond to unique tokens (i.e.\ symbols or indexes used for the input/output layers of NMT) while the less frequent ones are decomposed in two or more subwords (which are then translated as a sequence).
\item[Compositionality:] unlike other compression schemes that convert words to one or more symbols, BPE generates symbols that are word fragments, thus enabling generalization when translating unseen words by combining the translations of the subwords composing them. 
\item[Unknown words:] as individual characters are part of the vocabulary of tokens, any word in the test data can be tokenized, in the worst case into the characters that compose it.  Only words with characters not seen in the training data cannot be represented.
\end{description}

The \emph{compositionality} of BPE has often been presented as its main merit, though not without caveats.  Sennrich et al.~\shortcite{sennrich-etal-2016-neural} claimed that BPE ``is based on the intuition that various word classes are translatable via smaller units than words'' and on the analogy with a human translator who can translate some words ``even if they are novel to him or her, based on a translation of known subword units such as morphemes or phonemes.''  Pointing to the difference with Huffman coding, the authors state that their ``symbol sequences are still interpretable as subword units'' which ``the network can generalize to translate and produce new words.'' 
Quantitatively, Sennrich et al.~\shortcite{sennrich-etal-2016-neural} found that among ``100 rare tokens (not among the 50,000 most frequent types) in the German training data, the majority of tokens are potentially translatable from English through smaller units,'' in particular the 21 compounds they observed.

It is not, however, entirely clear if subwords actually correspond to meaningful part of words, such as morphemes or components of compound words.  Sennrich et al.~\shortcite{sennrich-etal-2016-neural} acknowledged that ``not every segmentation we produce is transparent'' and that they ``expect no performance benefit from opaque segmentations,'' i.e.\ segmentations where the units do not have independent meanings.  For instance,
Sennrich et al.\ showed that BPE leads to nearly the same BLEU scores as an encoding that keeps the 50,000 most frequent words as unique symbols, and encodes all the others using bigrams of characters as symbols. 
The challenge is indeed for a neural network to learn the correct translation of a series of two or more meaningless subwords.  Still, as long as the characters are included in the vocabulary, BPE can tokenize any word, thus effectively solving the \emph{unknown word} problem -- a merit which is widely recognized.

The other main reason for the effectiveness of BPE is the central role that \emph{token frequency} plays in the construction of the vocabulary, hence in deciding when to segment a word or not.  BPE uses fewer symbols to encode frequent words than less frequent ones, and a sizable part of a BPE vocabulary is actually made of entire words (see Figure~\ref{fig:symbols-per-word} in Section~\ref{sec:huf-vs-bpe} below).  This means that a large proportion of the tokens in the data are encoded as individual symbols, and only a smaller proportion are segmented into subwords.  For instance, Kudo~\shortcite{kudo-2018-subword} recognize that ``an advantage of BPE ..\ is that it can effectively balance the vocabulary size .. and the number of tokens required to encode a sentence'', because when applying BPE ``common words remain as unique symbols.''  In other words, BPE is effective because it ``keeps the most frequent words intact while splitting the rare ones into multiple tokens'' \cite{provilkov-etal-2020-bpe}.  


\section{Subword Tokenization based on Huffman Coding}
\label{sec:huffman}

We now introduce an alternative subword tokenization method which decouples compositionality from frequency, and implements only the second aspect.  This method will enable us to understand which of these aspects has the largest impact on the performance of NMT.  Just as BPE was originally inspired by a text compression algorithm, we transform here input and output texts into series of symbols using an adaptation of Huffman's \shortcite{huffman1952method} frequency-based compression algorithm.

\subsection{Overview}

In order to use the Huffman coding, all source and target sentences are processed as follows:

\begin{enumerate} \setlength{\itemsep}{0pt} 

\item Tokenize each sentence into words using the Moses tokenizer \cite{koehn-etal-2007-moses}, 
and apply truecasing to the words.\footnote{See \href{http://www2.statmt.org/moses/?n=Moses.Baseline}{www2.statmt.org/moses/?n=Moses.Baseline}.}

\item For each language, count the number of occurrences of each word, sort them in decreasing order, and build a Huffman tree with \textit{n} symbols using the algorithm given below. 

\item Save the `word'$\leftrightarrow$`code' mappings resulting from the tree, for each language, where the codes are made of one or more among the \textit{n} allowed symbols.\footnote{Technically, in our implementation, symbols are drawn from Unicode's range of CJK Unicode Ideographs \cite[Ch.~18]{unicode} of which nearly 100,000 code points are defined, starting at code point 0x4E00.  This offers a displayable textual representation of symbols, with no control codes that may be wrongly interpreted by the NMT system.}

\item Encode the train and test sentences, replacing each token by its symbolic counterpart.  Separate tokens with the Unicode symbol for space (code point 0x2420).

\item Split all the Huffman codes into symbols and separate them with white spaces. This allows to use NMT directly on the resulting text files, processing each symbol as an individual token, similarly to any tokenized input/output.  The vocabulary size is thus the number of symbols used to build the Huffman trees plus the separator.

\item Train NMT using the encoded parallel data. 

\item Encode the test data.  If words unseen during training appear on the source side, mark them with a special ``unknown'' symbol.  

\item Translate the encoded test data with NMT into encoded output.

\item Detokenize the NMT output by joining the symbols that are not separated by the 0x2420 separator symbol.  Then, decode the symbols using the `word'$\leftrightarrow$`code' mappings.  Sequences of symbols that have not been seen at training time, and are therefore absent from the mapping, are ignored.

\item Score the translated text by comparing it to the reference translation using BLEU, ChrF and COMET (see Section~\ref{sec:data-systems}).

\end{enumerate} 

\subsection{Building Huffman Trees}

Huffman trees can be built in several ways, resulting in different patterns of depth imbalance, which can be optimized depending on the relative frequencies of items to encode.  For all patterns, frequent items are placed higher in the tree, so that they are coded with fewer symbols.  We adapt the method as follows, being closest, although not identical to Chitnis and DeNero's \shortcite{chitnis-denero-2015-variable} ``Repeat-Symbol'' variant, with the main exception that we encode all tokens.

\begin{figure*}[ht]
 \centering
 \includegraphics[width=0.87\linewidth]{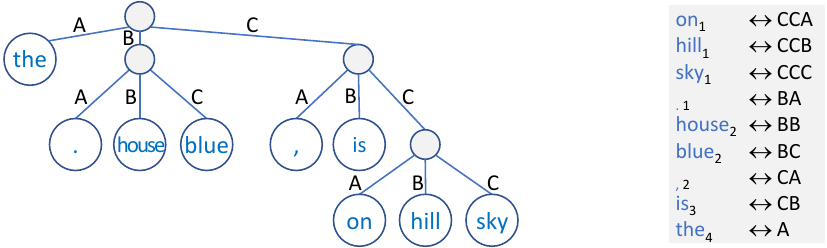}
 \caption{Ternary Huffman tree illustrating our approach.  The tree is built with Algorithm~\ref{alg:bhrigu} from word frequencies, shown as indices in the mapping (right), based on the following text: ``the house is on the hill, the house is blue, the sky is blue.''}\label{fig:huffman-binary}
\end{figure*}

We use $n$-ary Huffman trees, which are unbalanced trees in which the tokens to code appear on the leaves, and the paths leading to them constitute their encoded representations, i.e.\ the sequences of symbols on the branches.  This is illustrated in Figure~\ref{fig:huffman-binary} for a ternary tree with three symbols, which encodes eight word types based on their frequencies in a toy example.

\begin{algorithm}[h]
\rule{0.42\textwidth}{0.3pt}\\
 \KwData{Word frequencies $F$: $\{(w_i$, $f_i$), $\ldots\}$, Priority queue $H$: $\{(node_i$, $score_i$), $\ldots\}$ sorted by increasing scores, Number of symbols: $n$}
 \KwResult{Huffman tree}
 \ForEach{($w_i$, $f_i$) $\in$ F}{
  Create $node_i$ with key $w_i$ and score $f_i$\;
  Add $node_i$ to $H$\;
 }
 \While{length($H$) $> 1$}{
 $L$ $\gets$ empty list of nodes\;
 $S$ $\gets$ 0\;
\For{i $\gets$ 0 \KwTo $n$}{
\eIf{$H$ = $\emptyset$}{break\;}{
 Pop ($node_i$, $score_i$) from $H$\;
 Append ($node_i$, $score_i$) to $L$\;
 Add $score_i$ to $S$\;}
 }
 Create new node $N$ = (`None', $S$)\;
 \ForEach{node $\in$ $L$}{
 Add $node$ to $N$'s children\;}
 Push $N$ to $H$\;
 } 
 \rule{0.42\textwidth}{0.3pt}
 \vspace{.5em}
\caption{Construction of Huffman tree.}\label{alg:bhrigu} 
\end{algorithm}

\newpage
The number of coding symbols $n$ corresponds to the vocabulary size of the NMT system (the number of input units or indexes).  Each node has at most $n$ children, each one labeled with a symbol.  Each word type appearing on the source side of the training data (then, respectively, on the target side) is placed on a leaf of the tree, and the symbols on the path leading to it provide its representation with the new vocabulary of symbols.  For instance, if `the' is at the leaf stemming from the 10th branch of the root, it will be coded with symbol \#10, while if `control' can be reached through the 123rd then the 54th branch, it will be coded with two symbols, \#123\#54.  Whatever the value of $n \geq$ 2, a Huffman tree can encode an arbitrary large number of words, but the tree becomes deeper as $n$ decreases.

We use an open-source implementation of an algorithm building a Huffman tree.\footnote{Available at \href{https://github.com/bhrigu123/huffman-coding}{github.com/bhrigu123/huffman-coding} and explained in a blog entry \cite{bhrigu}.}  We have modified the code to make it applicable to words rather than to characters, and to generate a $n$-ary tree instead of a binary one, resulting in Algorithm~\ref{alg:bhrigu} above.  The key data structure is a priority queue with nodes and scores, always sorted by increasing scores, and initialized with the word types and their frequencies from the training data.  


Once Algorithm~\ref{alg:bhrigu} is run, each node of the resulting tree has at most \textit{n} children, therefore we can associate symbols to each of them, recursively doing the same operation for any node with the `None' label (i.e.\ not a leaf, which has a word label). At the end of this allocation, every node has a unique code of symbols, which is the concatenation of the symbols from the branches leading to it.  Leaves which are closer to the root have a shorter code than deeper leaves.  Our library\footnote{Available at \href{https://github.com/heig-iict-ida/huffman-tokenizer}{github.com/heig-iict-ida/huffman-tokenizer}.} supports large input texts, creates mappings between words and symbols, and allows encoding and decoding of texts.


\subsection{Properties of Our Method}

Prior to NMT, the codes produced by the Huffman algorithm are segmented into the symbols that compose them.  Therefore, the vocabulary size is $n$, the number of symbols.  In the Huffman tree, the most frequent words will appear as leaves close to the root.  Therefore, in the resulting mapping, \emph{the most frequent words will be represented with a single symbol}, and less frequent ones will use more symbols, which is considered as one of the advantages of BPE (see Section~\ref{sec:3properties}).  

However, unlike BPE, we do not segment words into subwords, hence \emph{we do not take into account the compositionality of subwords}, in the sense that words starting with a similar prefix are not encoded into Huffman codes starting with similar symbols.  For instance, the compositionality of BPE means that if `restor', `ing' and `ation' are subwords, then the NMT system can use knowledge about the translation of `restoring' to translate `restoration' (assuming they are tokenized as `restor' + `ing' and `restor' + `ation')  because both words share a common, meaningful prefix.  But if two Huffman codes share the same prefix, such as `\#10\#32' and `\#10\#76\#25', knowledge about the translation of `\#10' cannot be reused from one word to another, because the original words are unrelated.  This is why our study quantifies the utility of frequency alone, by separating it from compositionality. 

In addition, as the Huffman tree is built over words in the training data, it cannot encode unknown words in the test data, an effect that will be quantified below.

\section{Data and Systems}
\label{sec:data-systems}

We experiment with several language pairs featuring Czech, German, English, and French.  The training and test data come mostly from WMT 2014 \cite{bojar-etal-2014-findings} and WMT 2019 \cite{barrault-etal-2019-findings} and include also the JW300 data \cite{agic-vulic-2019-jw300}. The Czech-German data is shown in Table~\ref{tab:data1}, the English-German data in Table~\ref{tab:data2} and the English-French data in Table~\ref{tab:data3}.  We sample randomly from each subcorpus 0.1-0.2\% of sentences to serve as test data.  This particular split is made available with our library, for reproducibility.


\begin{table}[ht]
\begin{center}
\begin{tabular}{l|r}
\textbf{Data set} & \textbf{Number of lines} \\
\hline
News Commentary v14 & 172,995 \\
Europarl v9 & 556,182 \\
JW300 & 1,052,338 \\
Newstest 2019 & 1,997 \\
\hline
Total & 1,783,512 \\
Train / Test & 1,780,068 / 3,444 \\
\end{tabular}
\caption{Czech-German parallel data (non-empty lines).}\label{tab:data1}
\end{center}
\end{table}

\begin{table}[ht]
\begin{center}
\begin{tabular}{l|r}
\textbf{Data set} & \textbf{Number of lines} \\
\hline
Common Crawl & 2,399,123 \\
Europarl v7 & 1,911,843 \\
News Commentary v11 & 241,094 \\
\hline
Total & 4,552,060 \\
Train / Test & 4,547,445 / 4,615 \\
\end{tabular}
\caption{English-German parallel data (non-empty lines).}\label{tab:data2}
\end{center}
\end{table}

\begin{table}[ht]
\begin{center}
\begin{tabular}{l|r}
\textbf{Data set} & \textbf{Number of lines} \\
\hline
Common Crawl & 3,244,152 \\
Europarl v7 & 2,005,688 \\
\hline
Total & 5,249,840 \\
Train / Test & 5,245,392 / 4,448 \\
\end{tabular}
\caption{English-French parallel data (non-empty lines).}\label{tab:data3}
\end{center}
\end{table}

We use Transformer NMT models \cite{NIPS2017_3f5ee243} from the OpenNMT-py library \cite{klein-etal-2017-opennmt}
version 2.3.0.\footnote{\href{https://github.com/OpenNMT/OpenNMT-py}{github.com/OpenNMT/OpenNMT-py}}
We train the models for 150,000 steps, which takes about one day on two NVIDIA GeForce RTX 2080 Ti CPUs with 11 GB of memory.  The hyper-parameters of the models, generally the default ones, are given in Appendix~A. 
We evaluate the translation quality using the BLEU score~\cite{papineni-etal-2002-bleu}, the ChrF score~\cite{popovic-2015-chrf} as implemented by SacreBleu~\cite{post-2018-call}\footnote{BLEU score signature: \texttt{nrefs:1|case:mixed|eff:no|
tok:13a|smooth:exp|version:2.2.1}} and the COMET score~\cite{rei-etal-2020-comet}.
We compute the BLEU score obtained by each checkpoint (every 10,000 steps) on the test set, and select the best scoring checkpoint, on which we measure ChrF and COMET as well.


\section{NMT Using Huffman Coding: Results} 
\label{sec:scores}

In this section, we show that Huffman coding is a viable tokenization method, we study the impact of the number of available symbols on the translation quality, and compare the method with a purely frequential baseline.

We first investigate how translation quality changes according to the vocabulary size, which is the main hyper-parameter of the method.  If many symbols are available, then many frequent words will be encoded with a single symbol.  Conversely, fewer symbols result in most of the words being encoded with two or more symbols.  Figure~\ref{fig:symbols-per-word} below illustrates this property for Huffman codes with respect to BPE.

\begin{figure*}[ht]
 \centering
 \includegraphics[width=0.49\linewidth]{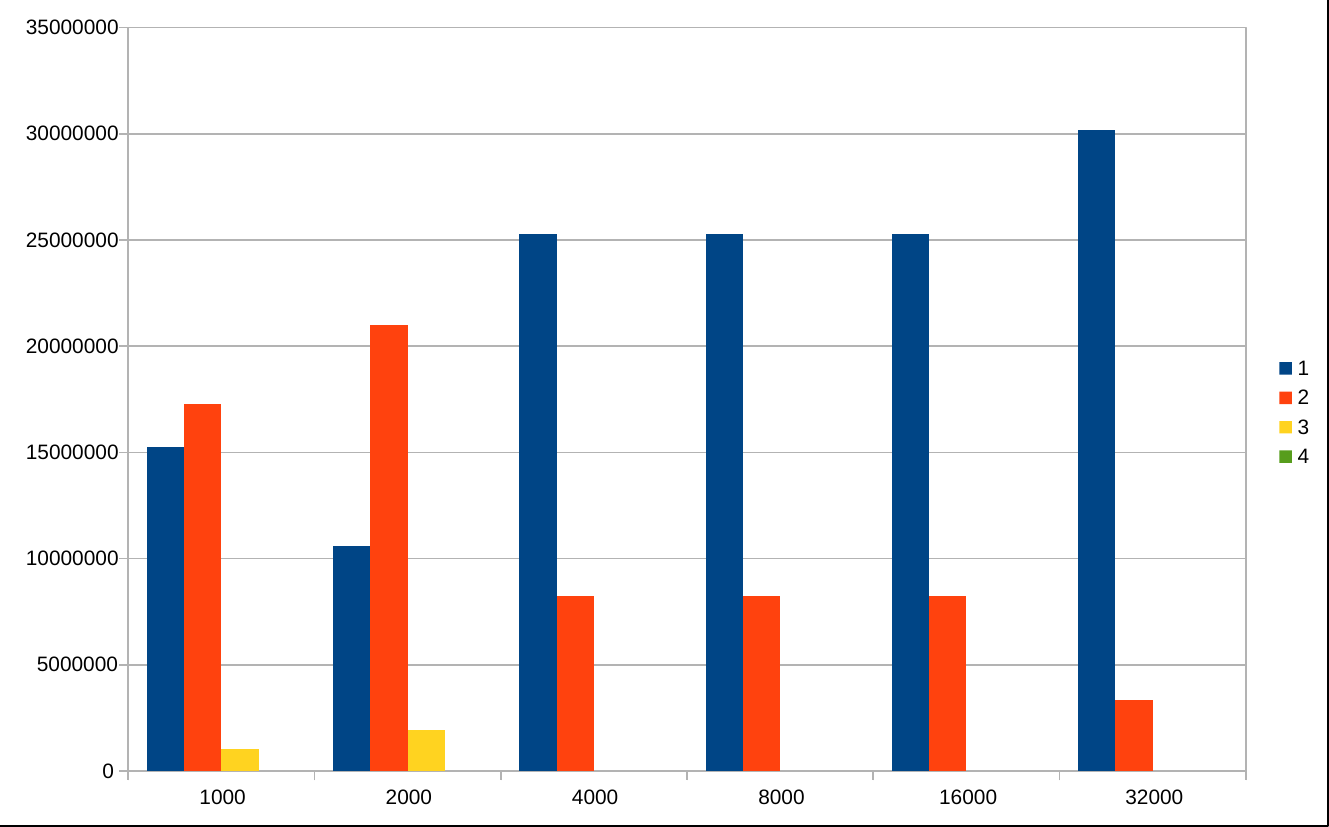}
 \includegraphics[width=0.49\linewidth]{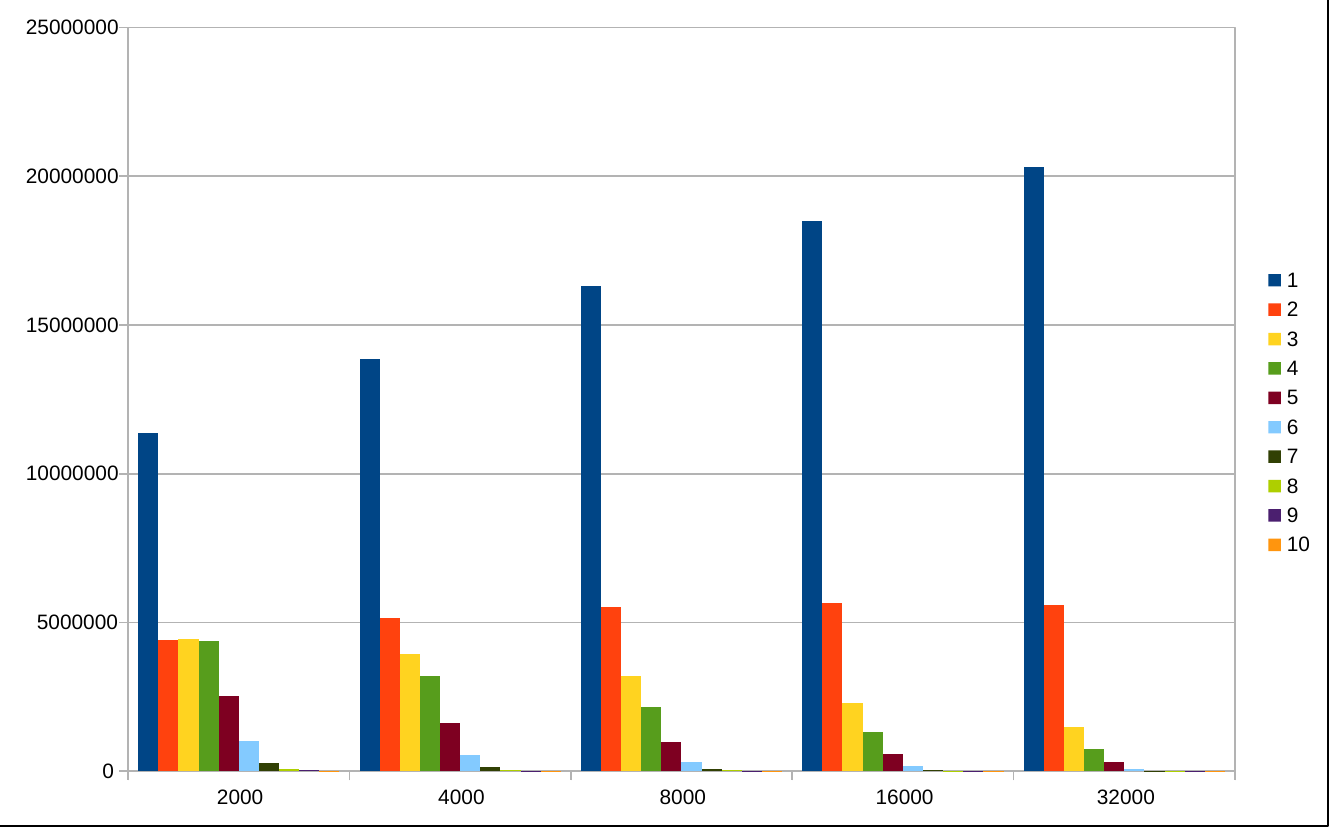}
 \caption{Histograms of the number of tokens from the CS data that are segmented into 1, 2, or more symbols, for Huffman coding (left) vs.\ BPE (right).  Six different vocabulary sizes are shown for Huffman coding (from 1k to 32k symbols) and five for BPE (from 2k to 32k merges).  While Huffman coding uses at most 4 symbols per token, BPE may use up to 10 subwords.}\label{fig:symbols-per-word}
\end{figure*}

The scores obtained with Huffman coding on CS-DE NMT with various numbers of symbols, shown in 
the first five lines of Table~\ref{tbl:huf-bpe-symbols},
demonstrate that the method is operational and that it benefits from an increasing number of symbols.  When the number of available symbols is very low, the effect on tokenization is closer to character-based translation, with the exception that some frequent words are still coded on one symbol with Huffman, while virtually all words contain two characters or more.  Not shown in the table, the BLEU score with 1,000 symbols is 19.6, which is very close to the BLEU score of a \emph{character-based Transformer} using a vocabulary of 485 characters, which is 19.4.  Our best scores, however, are found for higher vocabulary sizes, similar to those used with BPE (as discussed in Section~\ref{sec:huf-vs-bpe} below), which means we are conceptually closer to subwords than to character-based models. 

We studied the influence of several hyper-parameters on the CS-DE BLEU scores when using 1,000 symbols for Huffman coding.   As shown in Table~\ref{tbl:hyperparameters}, smaller embedding sizes (from 512 to 64) lead to substantially lower BLEU scores.  Increasing the number of Transformer layers from 8 to 20 appears to increase the scores, which is consistent for instance with the findings of Gupta et al.~\shortcite{Gupta2019} for character-based NMT.  However, the effect is not strong, and as the training costs increase substantially, we keep using 8-layer Transformers when comparing to BPE.  Finally, we note that the number of attention heads (8, 16, or 32) has almost no influence on scores. 

\begin{table}[ht]
\begin{center}
\begin{tabular}[c]{c|c|c|c}
     \textbf{Emb.\ size} & \textbf{Layers} & \textbf{Heads} & \textbf{BLEU} \\
     \hline
     512 & 8 & 8 & 19.6 \\
     256 & - & - & 17.6 \\
     128 & - & - & 14.4 \\
     64 & - & - & 9.7 \\
     \hline
     512 & 12 & 8 & 19.6 \\
     - & 16 & - & 21.4 \\
     - & 20 & - & 21.4 \\
     \hline
     512 & 8 & 16 & 19.3 \\
     - & - & 32 & 19.4
\end{tabular}
\end{center}
\caption{BLEU scores with 1000-symbol Huffman coding when varying the embedding size of the Transformer, the number of layers, and the number of attention heads (`-' means ``same as above'').}\label{tbl:hyperparameters}
\end{table}

We also compare Huffman coding with a baseline that simply keeps as symbols the most frequent tokens in the source and target training texts, all other parameters being equal.  Keeping 16k tokens leads to a BLEU score of 17.0 (compared to 22.3 for Huffman with 16k symbols) and keeping 32k tokens leads to 19.1 BLEU points (compared to 23.1 for Huffman with 32k symbols).  As the new scores are only 4--5 points lower, we conclude that focusing on the most frequent tokens preserves some effectiveness, especially as the vocabulary grows, but is limited by the fact that all other tokens are ignored.



\begin{table*}[t]
\begin{center}
\begin{tabular}{l|c|c|c|c|c|c|c|c|c|c}
\textbf{Lang.} & \textbf{Nb.\ of} & \multicolumn{3}{c|}{\bf BLEU} & \multicolumn{3}{c|}{\bf ChrF} & \multicolumn{3}{c}{\bf COMET} \\
\textbf{pair} & \textbf{symbols} & \textbf{Huffman} & \textbf{BPE} & \textbf{\%} & \textbf{Huffman} & \textbf{BPE} & \textbf{\%} & \textbf{Huffman} & \textbf{BPE} & \textbf{\%} \\ \hline
CS-DE 
      & 2k & 20.3 & 24.4 & 83.2 & 46.6 & 52.6 & 88.6 & 0.758 & 0.829 & 91.4\\
      & 4k & 20.9 & 24.8 & 84.3 & 47.2 & 53.2 & 88.7 & 0.762 & 0.833 & 91.4\\ 
      & 8k & 21.6 & 25.1 & 86.1 & 48.4 & 53.4 & 90.6 & 0.780 & 0.834 & 93.6\\ 
      & 16k & 22.3 & 24.8 & 89.9 & 49.3 & 53.3 & 92.5 & 0.791 & 0.830 & 95.2\\
      & 32k & 23.1 & 26.4 & 87.5 & 50.2 & 54.5 & 92.1 & 0.804 & 0.837 & 96.0\\ 
\hline
EN-DE & 8k & 19.5 & 22.4 & 87.1 & 46.4 & 49.7 & 93.4 & 0.709 & 0.769 & 92.2\\
      & 16k & 20.3 & 22.2 & 91.4 & 46.6 & 49.3 & 94.5 & 0.718 & 0.768 & 93.5\\
      & 32k & 19.8 & 22.5 & 88.0 & 46.9 & 49.5 & 94.7 & 0.712 & 0.772 & 92.2\\ 
\hline
EN-FR & 8k & 27.1 & 31.2 & 86.9 & 51.1 & 55.3 & 92.4 & 0.728 & 0.783 & 93.0\\
      & 16k & 27.6 & 30.9 & 89.3 & 51.8 & 55 & 94.2 & 0.739 & 0.781 & 94.6\\ 
      & 32k & 27.9 & 30.9 & 90.3 & 52.2 & 54.9 & 95.1  &0.746 & 0.784 & 95.1\\
\end{tabular}
\end{center}
\caption{Translation quality achieved by Huffman and BPE models with increasing numbers of symbols.}\label{tbl:huf-bpe-symbols}
\end{table*}

\section{Huffman Coding versus BPE}
\label{sec:huf-vs-bpe}

In order to compare BPE with Huffman models, we tokenize source and target sides jointly for each pair using BPE from the SentencePiece toolkit (see footnote~1) with an increasing number of merges: 2k, 4k, 8k, 16k, and 32k. 

We compare first the vocabularies resulting from BPE with those from Huffman coding in terms of the number of symbols per token.  The histograms of the numbers of tokens having respectively 1, 2, 3, and up to 10 symbols~/ subwords are shown in Figure~\ref{fig:symbols-per-word}, side-by-side for Huffman coding (left) and for BPE (right), for each vocabulary size.  As the vocabulary size (or number of symbols) grows, tokenization results become more similar across the two methods, with more than 3/4 of the tokens being kept as unique symbols.  While two is the maximum number of symbols per token for Huffman coding, by construction, we see that for BPE some tokens are segmented into 3 or more subwords (up to 10, although their number is too small to be seen in the figure).  These observations support our claim that Huffman coding captures similar frequency-related information as BPE, while by design it does not capture compositionality.

Turning now to NMT scores, Table~\ref{tbl:huf-bpe-symbols} compares those of BPE-based models with their Huffman counterpart for three language pairs and three metrics.  We observe that increasing the number of BPE merges has a positive but rather limited impact in this setting, with an improvement of only 2 BLEU points between the best 2k and 32k models.  On all language pairs, the Huffman and BPE scores become more similar as the numbers of symbols increase, as shown in the `\%' column that indicates the ratio between Huffman and BPE scores (with one exception, EN-DE with 32k symbols).  Beyond 8k symbols, our method obtains between 86.1\% and 91.4\% of the BLEU score of BPE for all language pairs, and even higher fractions for ChrF (between 90.6\% and 95.1\%) and COMET (between 92.2\% and 96.0\%).  Still, the BPE models always outperform their Huffman equivalent by 2-3 BLEU points all language pairs. 

We attribute these differences to the fact that Huffman coding relies on frequency only to select the number of subwords per token, and does not benefit from compositionality.  We interpret the results as a quantification of the importance of frequency vs.\ compositionality in subword tokenization, with a large part of the final performance coming from frequency and the remaining difference (between 4 and 14 percentage points depending on the metric) to compositionality and the capacity to deal with unknown words.   Another reason for the remaining difference is the fact that the BPE vocabulary is built jointly on the source and target data, unlike our method.  

Finally, unknown words are also likely to limit the performance of Huffman coding, although their number is very small in the test data.  There are 0.55\% unknown tokens in the CS source for CS-DE, 0.46\% in the EN source for EN-DE, and none in the EN source for EN-FR.  Interestingly, on the decoding side, the vast majority of symbol combinations generated by our NMT models correspond to actual leaves of Huffman trees: the percentages of unknown combinations of symbols among the total output tokens are respectively 0.07\%, 0.04\% and 0.02\% for CS-DE, EN-DE, and EN-FR.  Such combinations cannot be decoded and are therefore skipped.

\section{Conclusion}

In this paper, we have presented an original method for text tokenization, which exploits the text compression property of Huffman trees, and therefore takes into account the frequencies of subwords, but does not rely on their compositionality.  We have framed these notions and, based on the comparison of scores obtained with Huffman coding with those obtained with BPE, we have defended the claim that most of the gains brought by BPE are due to the appropriate consideration of subword frequency, and comparatively much less to compositionality.  These results tend to downplay the importance of compositionality, which is often mentioned as an advantage of BPE, and contribute to the understanding of the remarkable effectiveness of this method.

We hypothesize that text compression methods might provide inspiration, in the future, for even more effective tokenization methods, given that the state-of-the-art in compression has made significant progress since BPE.  Especially, Prediction by Partial Matching seems a promising candidate, but awaits a principled solution to relate tokens to coding symbols.

\section*{Acknowledgments}

We are grateful for the support received from Armasuisse (UNISUB projet: Unsupervised NMT with Innovative Multilingual Subword Models) and from the Swiss National Science Foundation (DOMAT project: On-demand Knowledge for Document-level Machine Translation, n.\ 175693).  We thank the four anonymous EAMT reviewers for their suggestions, and Mr.\ Bhrigu Srivastava for sharing his implementation of Huffman coding.


\bibliography{eamt23}

\begin{thebibliography}{}

\bibitem[\protect\citename{Agi{\'c} and Vuli{\'c}}2019]{agic-vulic-2019-jw300}
Agi{\'c}, {\v{Z}}eljko and Ivan Vuli{\'c}.
\newblock 2019.
\newblock {JW}300: A wide-coverage parallel corpus for low-resource languages.
\newblock In {\em Proceedings of the 57th Annual Meeting of the Association for
  Computational Linguistics}, pages 3204--3210.

\bibitem[\protect\citename{Amrhein and
  Sennrich}2020]{amrhein-sennrich-2020-romanization}
Amrhein, Chantal and Rico Sennrich.
\newblock 2020.
\newblock On {R}omanization for model transfer between scripts in neural
  machine translation.
\newblock In {\em Findings of the Association for Computational Linguistics:
  EMNLP 2020}, pages 2461--2469.

\bibitem[\protect\citename{Barrault \bgroup et al.\egroup
  }2019]{barrault-etal-2019-findings}
Barrault, Lo{\"\i}c, Ond{\v{r}}ej Bojar, Marta~R. Costa-juss{\`a}, Christian
  Federmann, Mark Fishel, Yvette Graham, Barry Haddow, Matthias Huck, Philipp
  Koehn, Shervin Malmasi, Christof Monz, Mathias M{\"u}ller, Santanu Pal, Matt
  Post, and Marcos Zampieri.
\newblock 2019.
\newblock Findings of the 2019 {C}onference on {M}achine {T}ranslation
  ({WMT}19).
\newblock In {\em Proceedings of the Fourth Conference on Machine Translation},
  pages 1--61.

\bibitem[\protect\citename{Bojar \bgroup et al.\egroup
  }2014]{bojar-etal-2014-findings}
Bojar, Ond{\v{r}}ej, Christian Buck, Christian Federmann, Barry Haddow, Philipp
  Koehn, Johannes Leveling, Christof Monz, Pavel Pecina, Matt Post, Herve
  Saint-Amand, Radu Soricut, Lucia Specia, and Ale{\v{s}} Tamchyna.
\newblock 2014.
\newblock Findings of the 2014 {W}orkshop on {S}tatistical {M}achine
  {T}ranslation.
\newblock In {\em Proceedings of the Ninth Workshop on Statistical Machine
  Translation}, pages 12--58.

\bibitem[\protect\citename{Bojar \bgroup et al.\egroup
  }2015]{bojar-etal-2015-findings}
Bojar, Ond{\v{r}}ej, Rajen Chatterjee, Christian Federmann, Barry Haddow,
  Matthias Huck, Chris Hokamp, Philipp Koehn, Varvara Logacheva, Christof Monz,
  Matteo Negri, Matt Post, Carolina Scarton, Lucia Specia, and Marco Turchi.
\newblock 2015.
\newblock Findings of the 2015 {W}orkshop on {S}tatistical {M}achine
  {T}ranslation.
\newblock In {\em Proceedings of the Tenth Workshop on Statistical Machine
  Translation}, pages 1--46.

\bibitem[\protect\citename{Cherry \bgroup et al.\egroup
  }2018]{cherry-etal-2018-revisiting}
Cherry, Colin, George Foster, Ankur Bapna, Orhan Firat, and Wolfgang Macherey.
\newblock 2018.
\newblock Revisiting character-based neural machine translation with capacity
  and compression.
\newblock In {\em Proceedings of the 2018 Conference on Empirical Methods in
  Natural Language Processing}, pages 4295--4305.

\bibitem[\protect\citename{Chitnis and
  DeNero}2015]{chitnis-denero-2015-variable}
Chitnis, Rohan and John DeNero.
\newblock 2015.
\newblock Variable-length word encodings for neural translation models.
\newblock In {\em Proceedings of the 2015 Conference on Empirical Methods in
  Natural Language Processing}, pages 2088--2093.

\bibitem[\protect\citename{Denkowski and
  Neubig}2017]{denkowski-neubig-2017-stronger}
Denkowski, Michael and Graham Neubig.
\newblock 2017.
\newblock Stronger baselines for trustable results in neural machine
  translation.
\newblock In {\em Proceedings of the First Workshop on Neural Machine
  Translation}, pages 18--27.

\bibitem[\protect\citename{Gage}1994]{Gage1994ANA}
Gage, Philip.
\newblock 1994.
\newblock A new algorithm for data compression.
\newblock {\em The C Users Journal archive}, 12:23--38.

\bibitem[\protect\citename{Gall{\'e}}2019]{galle-2019-investigating}
Gall{\'e}, Matthias.
\newblock 2019.
\newblock Investigating the effectiveness of {BPE}: The power of shorter
  sequences.
\newblock In {\em Proceedings of the 2019 Conference on Empirical Methods in
  Natural Language Processing and the 9th International Joint Conference on
  Natural Language Processing (EMNLP-IJCNLP)}, pages 1375--1381.

\bibitem[\protect\citename{Gupta \bgroup et al.\egroup }2019]{Gupta2019}
Gupta, Rohit, Laurent Besacier, Marc Dymetman, and Matthias Gallé.
\newblock 2019.
\newblock Character-based {NMT} with {T}ransformer.

\bibitem[\protect\citename{Huffman}1952]{huffman1952method}
Huffman, David~A.
\newblock 1952.
\newblock A method for the construction of minimum-redundancy codes.
\newblock {\em Proceedings of the IRE}, 40(9):1098--1101.

\bibitem[\protect\citename{Jean \bgroup et al.\egroup
  }2015]{jean-etal-2015-using}
Jean, S{\'e}bastien, Kyunghyun Cho, Roland Memisevic, and Yoshua Bengio.
\newblock 2015.
\newblock On using very large target vocabulary for neural machine translation.
\newblock In {\em Proceedings of the 53rd Annual Meeting of the Association for
  Computational Linguistics}, pages 1--10.

\bibitem[\protect\citename{Klein \bgroup et al.\egroup
  }2017]{klein-etal-2017-opennmt}
Klein, Guillaume, Yoon Kim, Yuntian Deng, Jean Senellart, and Alexander Rush.
\newblock 2017.
\newblock {O}pen{NMT}: Open-source toolkit for neural machine translation.
\newblock In {\em Proceedings of {ACL} 2017, System Demonstrations}, pages
  67--72.

\bibitem[\protect\citename{Koehn \bgroup et al.\egroup
  }2007]{koehn-etal-2007-moses}
Koehn, Philipp, Hieu Hoang, Alexandra Birch, Chris Callison-Burch, Marcello
  Federico, Nicola Bertoldi, Brooke Cowan, Wade Shen, Christine Moran, Richard
  Zens, Chris Dyer, Ond{\v{r}}ej Bojar, Alexandra Constantin, and Evan Herbst.
\newblock 2007.
\newblock {M}oses: Open source toolkit for statistical machine translation.
\newblock In {\em Proceedings of the 45th Annual Meeting of the Association for
  Computational Linguistics, Demo and Poster Sessions}, pages 177--180.

\bibitem[\protect\citename{Kudo and
  Richardson}2018]{kudo-richardson-2018-sentencepiece}
Kudo, Taku and John Richardson.
\newblock 2018.
\newblock {S}entence{P}iece: A simple and language independent subword
  tokenizer and detokenizer for neural text processing.
\newblock In {\em Proceedings of the 2018 Conference on Empirical Methods in
  Natural Language Processing: System Demonstrations}, pages 66--71.

\bibitem[\protect\citename{Kudo}2018]{kudo-2018-subword}
Kudo, Taku.
\newblock 2018.
\newblock Subword regularization: Improving neural network translation models
  with multiple subword candidates.
\newblock In {\em Proceedings of the 56th Annual Meeting of the Association for
  Computational Linguistics}, pages 66--75.

\bibitem[\protect\citename{Lee \bgroup et al.\egroup }2017]{lee-character}
Lee, Jason, Kyunghyun Cho, and Thomas Hofmann.
\newblock 2017.
\newblock Fully character-level neural machine translation without explicit
  segmentation.
\newblock {\em Transactions of the Association for Computational Linguistics},
  5:365--378.

\bibitem[\protect\citename{Libovick{\'y} \bgroup et al.\egroup
  }2022]{libovicky-etal-2022-dont}
Libovick{\'y}, Jind{\v{r}}ich, Helmut Schmid, and Alexander Fraser.
\newblock 2022.
\newblock Why don{'}t people use character-level machine translation?
\newblock In {\em Findings of the Association for Computational Linguistics:
  ACL 2022}, pages 2470--2485.

\bibitem[\protect\citename{Luong and
  Manning}2016]{luong-manning-2016-achieving}
Luong, Minh-Thang and Christopher~D. Manning.
\newblock 2016.
\newblock Achieving open vocabulary neural machine translation with hybrid
  word-character models.
\newblock In {\em Proceedings of the 54th Annual Meeting of the Association for
  Computational Linguistics}, pages 1054--1063.

\bibitem[\protect\citename{Luong \bgroup et al.\egroup
  }2015]{luong-etal-2015-addressing}
Luong, Thang, Ilya Sutskever, Quoc Le, Oriol Vinyals, and Wojciech Zaremba.
\newblock 2015.
\newblock Addressing the rare word problem in neural machine translation.
\newblock In {\em Proceedings of the 53rd Annual Meeting of the Association for
  Computational Linguistics}, pages 11--19.

\bibitem[\protect\citename{Papineni \bgroup et al.\egroup
  }2002]{papineni-etal-2002-bleu}
Papineni, Kishore, Salim Roukos, Todd Ward, and Wei-Jing Zhu.
\newblock 2002.
\newblock {BLEU}: a method for automatic evaluation of machine translation.
\newblock In {\em Proceedings of the 40th Annual Meeting of the Association for
  Computational Linguistics}, pages 311--318.

\bibitem[\protect\citename{Patil \bgroup et al.\egroup
  }2022]{patil-etal-2022-overlap}
Patil, Vaidehi, Partha Talukdar, and Sunita Sarawagi.
\newblock 2022.
\newblock Overlap-based vocabulary generation improves cross-lingual transfer
  among related languages.
\newblock In {\em Proceedings of the 60th Annual Meeting of the Association for
  Computational Linguistics}, pages 219--233.

\bibitem[\protect\citename{Popovi{\'c}}2015]{popovic-2015-chrf}
Popovi{\'c}, Maja.
\newblock 2015.
\newblock chr{F}: character n-gram {F}-score for automatic {MT} evaluation.
\newblock In {\em Proceedings of the Tenth Workshop on Statistical Machine
  Translation}, pages 392--395.

\bibitem[\protect\citename{Post}2018]{post-2018-call}
Post, Matt.
\newblock 2018.
\newblock A call for clarity in reporting {BLEU} scores.
\newblock In {\em Proceedings of the Third Conference on Machine Translation:
  Research Papers}, pages 186--191.

\bibitem[\protect\citename{Provilkov \bgroup et al.\egroup
  }2020]{provilkov-etal-2020-bpe}
Provilkov, Ivan, Dmitrii Emelianenko, and Elena Voita.
\newblock 2020.
\newblock {BPE}-dropout: Simple and effective subword regularization.
\newblock In {\em Proceedings of the 58th Annual Meeting of the Association for
  Computational Linguistics}, pages 1882--1892.

\bibitem[\protect\citename{Rei \bgroup et al.\egroup
  }2020]{rei-etal-2020-comet}
Rei, Ricardo, Craig Stewart, Ana~C Farinha, and Alon Lavie.
\newblock 2020.
\newblock {COMET}: A neural framework for {MT} evaluation.
\newblock In {\em Proceedings of the 2020 Conference on Empirical Methods in
  Natural Language Processing (EMNLP)}, pages 2685--2702.

\bibitem[\protect\citename{Schuster and Nakajima}2012]{Schuster-Nakajima-2012}
Schuster, Mike and Kaisuke Nakajima.
\newblock 2012.
\newblock Japanese and {K}orean voice search.
\newblock In {\em 2012 IEEE International Conference on Acoustics, Speech and
  Signal Processing (ICASSP)}, pages 5149--5152.

\bibitem[\protect\citename{Sennrich \bgroup et al.\egroup
  }2016]{sennrich-etal-2016-neural}
Sennrich, Rico, Barry Haddow, and Alexandra Birch.
\newblock 2016.
\newblock Neural machine translation of rare words with subword units.
\newblock In {\em Proceedings of the 54th Annual Meeting of the Association for
  Computational Linguistics (Volume 1: Long Papers)}, pages 1715--1725.

\bibitem[\protect\citename{Srivastava}2017]{bhrigu}
Srivastava, Bhrigu.
\newblock 2017.
\newblock Huffman coding {P}ython implementation.
\newblock {\em Personal Blog}.

\bibitem[\protect\citename{Unicode~Consortium}2022]{unicode}
Unicode~Consortium, The.
\newblock 2022.
\newblock {\em The Unicode Standard, Version 15.0.0}.
\newblock The Unicode Consortium.

\bibitem[\protect\citename{Vaswani \bgroup et al.\egroup
  }2017]{NIPS2017_3f5ee243}
Vaswani, Ashish, Noam Shazeer, Niki Parmar, Jakob Uszkoreit, Llion Jones,
  Aidan~N Gomez, {\L}ukasz Kaiser, and Illia Polosukhin.
\newblock 2017.
\newblock Attention is all you need.
\newblock In Guyon, I., U.~Von Luxburg, S.~Bengio, H.~Wallach, R.~Fergus,
  S.~Vishwanathan, and R.~Garnett, editors, {\em Advances in Neural Information
  Processing Systems}, volume~30.

\bibitem[\protect\citename{Vernikos and
  Popescu-Belis}2021]{vernikos-popescu-belis-2021-subword-mapping}
Vernikos, Giorgos and Andrei Popescu-Belis.
\newblock 2021.
\newblock Subword mapping and anchoring across languages.
\newblock In {\em Findings of the Association for Computational Linguistics:
  EMNLP 2021}, pages 2633--2647.

\bibitem[\protect\citename{Wu \bgroup et al.\egroup }2016]{Wu2016GooglesNM}
Wu, Yonghui, Mike Schuster, Z.~Chen, Quoc~V. Le, Mohammad Norouzi, Wolfgang
  Macherey, Maxim Krikun, Yuan Cao, Qin Gao, Klaus Macherey, Jeff Klingner,
  Apurva Shah, Melvin Johnson, Xiaobing Liu, Lukasz Kaiser, Stephan Gouws,
  Yoshikiyo Kato, Taku Kudo, Hideto Kazawa, Keith Stevens, George Kurian,
  Nishant Patil, Wei Wang, Cliff Young, Jason~R. Smith, Jason Riesa, Alex
  Rudnick, Oriol Vinyals, Gregory~S. Corrado, Macduff Hughes, and Jeffrey Dean.
\newblock 2016.
\newblock Google's neural machine translation system: Bridging the gap between
  human and machine translation.
\newblock {\em ArXiv}, abs/1609.08144.

\end{thebibliography}
\bibliographystyle{eamt23}

\vspace{1em}
\appendix
\section*{Appendix A. Parameters of OpenNMT-py} \label{sec:app_a}

The hyper-parameters we used for our experiments with OpenNMT-py are the following ones:

\begin{itemize}
\setlength\itemsep{-1mm}
\item Number of layers: 8
\item Number of heads: 8
\item Embedding size: 512
\item Transformer feed-forward size: 2048
\item Batch size: 2,000 tokens
\item Optimizer: Adam
\item Learning rate factor: 2.0 
\item Warmup steps: 8,000
\item Dropout rate: 0.1
\end{itemize}

\end{document}